\documentclass[conference]{IEEEtran}
\IEEEoverridecommandlockouts
\usepackage{cite}
\usepackage{amsmath,amssymb,amsfonts}
\usepackage{algorithmic}
\usepackage{graphicx}
\usepackage{textcomp}
\usepackage{xcolor}
\usepackage{multirow}
\usepackage{array}
\def\BibTeX{{\rm B\kern-.05em{\sc i\kern-.025em b}\kern-.08em
    T\kern-.1667em\lower.7ex\hbox{E}\kern-.125emX}}
\begin{document}

\title{Evaluating the Significance of Outdoor Advertising from Driver's Perspective Using Computer Vision
}

\author{\IEEEauthorblockN{1\textsuperscript{st} Zuzana Černeková}
\IEEEauthorblockA{\textit{Faculty of Mathematics Physics and informatics} \\
\textit{Comenius University}\\
Bratislava, Slovakia \\
zuzana.cernekova@fmph.uniba.sk \\
0000-0002-7617-4192}
\and
\IEEEauthorblockN{2\textsuperscript{nd} Zuzana Berger Haladová}
\IEEEauthorblockA{\textit{Faculty of Mathematics Physics and informatics} \\
\textit{Comenius University}\\
Bratislava, Slovakia \\
zuzana.bergerhaladova@fmph.uniba.sk \\
0000-0002-5947-8063}
\and
\IEEEauthorblockN{3\textsuperscript{rd} Ján Špirka}
\IEEEauthorblockA{\textit{Faculty of Mathematics Physics and informatics} \\
\textit{Comenius University}\\
Bratislava, Slovakia \\
janspirka98@gmail.com}
\and
\IEEEauthorblockN{4\textsuperscript{th} Viktor Kocur}
\IEEEauthorblockA{\textit{Faculty of Mathematics Physics and informatics} \\
\textit{Comenius University}\\
Bratislava, Slovakia \\
viktor.kocur@fmph.uniba.sk \\
0000-0001-8752-2685}
}

\IEEEoverridecommandlockouts
\IEEEpubid{\makebox[\columnwidth]{979-8-3503-4353-3/23/\$31.00~\copyright2023 IEEE\hfill} \hspace{\columnsep}\makebox[\columnwidth]{ }}

\maketitle
\IEEEpubidadjcol
\begin{abstract}

Outdoor advertising, such as roadside billboards, plays a significant role in marketing campaigns but can also be a distraction for drivers, potentially leading to accidents. In this study, we propose a pipeline for evaluating the significance of roadside billboards in videos captured from a driver's perspective. We have collected and annotated a new BillboardLamac dataset, comprising eight videos captured by drivers driving through a predefined path wearing eye-tracking devices. The dataset includes annotations of billboards, including 154 unique IDs and 155 thousand bounding boxes, as well as eye fixation data. We evaluate various object tracking methods in combination with a YOLOv8 detector to identify billboard advertisements with the best approach achieving 38.5 HOTA on BillboardLamac. Additionally, we train a random forest classifier to classify billboards into three classes based on the length of driver fixations achieving 75.8\% test accuracy. An analysis of the trained classifier reveals that the duration of billboard visibility, its saliency, and size are the most influential features when assessing billboard significance. 
\end{abstract}


\section{Introduction}
Acquisition advertising is frequently employed as part of a marketing strategy to increase brand awareness or support ongoing campaigns. Businesses utilize various advertising platforms such as television, radio, and social networks. One common outdoor advertising method is roadside advertising (billboards), positioned along roads and highways to attract the attention of drivers and other commuters. Additionally, advertisements can be found on buildings, fences, bus stops, and even vehicles, creating a visual presence in road traffic. Unfortunately, these ads are often considered visual pollution on the roads.

The impact of advertising on individuals depends on various factors, including the type of advertising, strategic placement, and the content itself. Billboards undoubtedly capture the attention of drivers, benefiting marketing efforts. However, it is important to acknowledge that even a momentary loss of driver attention can result in traffic accidents with severe consequences. Several studies, including our work, deal with the issue of advertisement safety. 

In this paper, we have created a pipeline for the estimation of billboard advertisement's significance. The pipeline consists of billboard detection and tracking and classification of the billboard significance which is evaluated by data collected in our user study utilizing eyetracking technology.

The paper is organized as follows. In the second section, we discuss the previous works on billboard advertisements' characteristics and significance. In the third section, we describe our dataset BillboardLamac. The fourth section is dedicated to the billboard tracking methods description and evaluation. The fifth section discusses the classification of the billboard's significance.

\section{Billboards}
The following section explores the previous works on the effectiveness of electronic billboards and traditional billboards, considering factors such as gaze duration, placement, and content impact on driver attention and road safety.

Several studies indicate that electronic billboards achieve longer gaze duration compared to traditional non-illuminated advertising \cite{Oviedo19,Beijer04,Brome21}. This type of advertisement is the most dangerous due to the average duration of fixations \cite{Yellappan16,Smiley05}.

The placement of advertising significantly influences its reach. Advertisements located on the driver's side of the road recorded more fixations than those on the opposite side \cite{Costa19}. The assumption that the height of the advertisement placement prolongs fixation duration was negatively confirmed in studies \cite{Costa19, Crundall06}. The authors found that drivers looked more at advertisements at road level than those elevated three or more meters above the ground. An exception occurs when drivers deliberately seek out advertisements, as they tend to look higher. Furthermore, some studies considered the angle of the advertising surface \cite{Zalesinska18}, the complexity of the traffic section, the type of environment, and its complexity \cite{Mollu18,Costa19}. It was found that all these elements to some extent influence the visibility of advertisements.

The content of the advertisement is another important characteristic. Studies \cite{Harasimczuk21, Meuleners20} confirm the assumption that advertisements with longer text achieve longer fixations. This also applies to advertisements with a sexual undertone \cite{Maliszewski19} and advertisements featuring human beings \cite{Tarnowski17}. Moreover, the more complex the advertisement, the longer the driver's gaze on it \cite{Marciano17}. The content of the advertisement that elicits negative emotions in drivers was also examined \cite{Chan13}. It was found that such content also leads to longer fixations and, at the same time, reduces speed. Additionally, questionnaires filled out by drivers after their drive showed that negative advertisements are the most memorable.

\section{Dataset}
\begin{figure}
    \centering
    \includegraphics[width=0.48\textwidth]{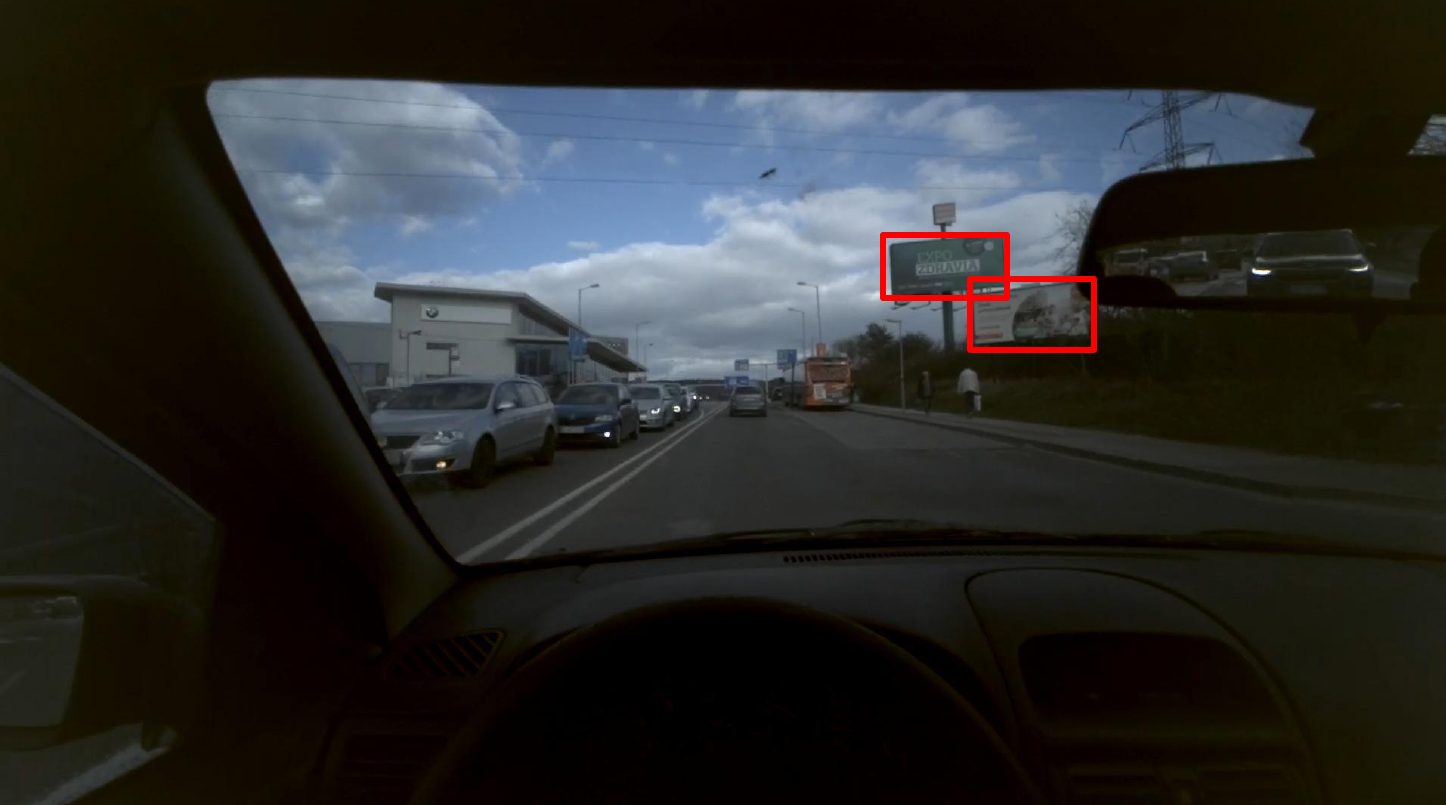}
    \caption{A frame from the dataset BillboardLamac with annotated billboards (ID 67 and 68) and their corresponding bounding boxes.}
    \label{fig:anot}
\end{figure}

\begin{figure}
    \centering
    \includegraphics[width=0.48\textwidth]{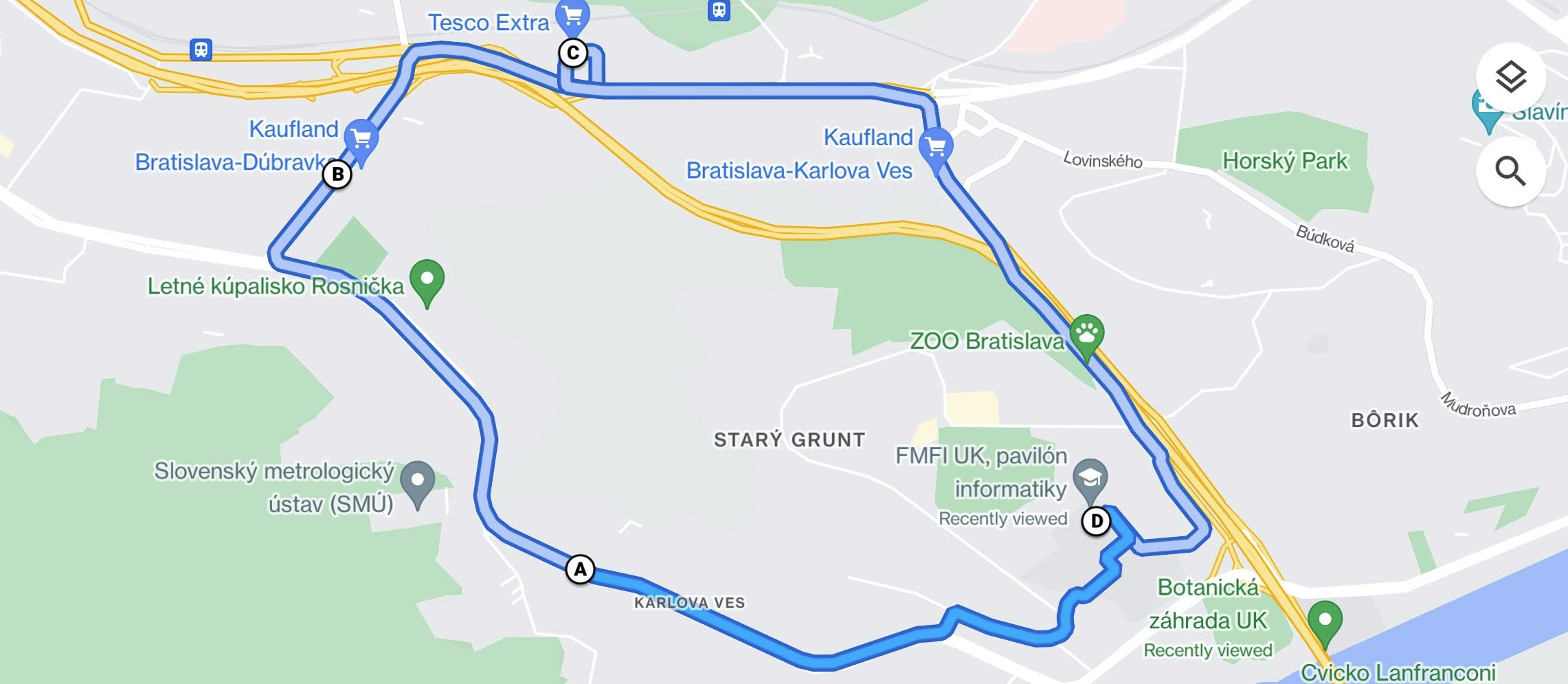}
    \caption{Map of the round trip taken by the participants of the user study displayed in Google Maps.}
    \label{fig:map}
\end{figure}

To test our method for billboard detection, tracking, and saliency evaluation we have created a new BillboardLamac dataset which consists of 8 videos captured by an eyetracker worn by 8 different drivers on the same round trip in Lamač, Bratislava, Slovakia. A map of the round trip taken by the participants of the user study is shown in Fig. \ref{fig:map}. We have annotated the billboards in the videos, for each billboard on the roundtrip we have a unique ID (which is consistent through all the videos), and the bounding box for each frame where it is visible. There are 145 unique billboards annotated with 155 912 bounding boxes in 8 videos of an average length of 24 minutes. An example of an image from the dataset with displayed annotations can be seen in Fig. \ref{fig:anot}.
The dataset is available on demand and includes videos from the eyetracker, a text file with bounding boxes and billboards' IDs with the number of corresponding frames of the video, and the text file with positions of the eye fixations for each frame of the video. 

\subsection{Eyetracking}
\begin{figure}
    \centering
    \includegraphics[width=0.48\textwidth]{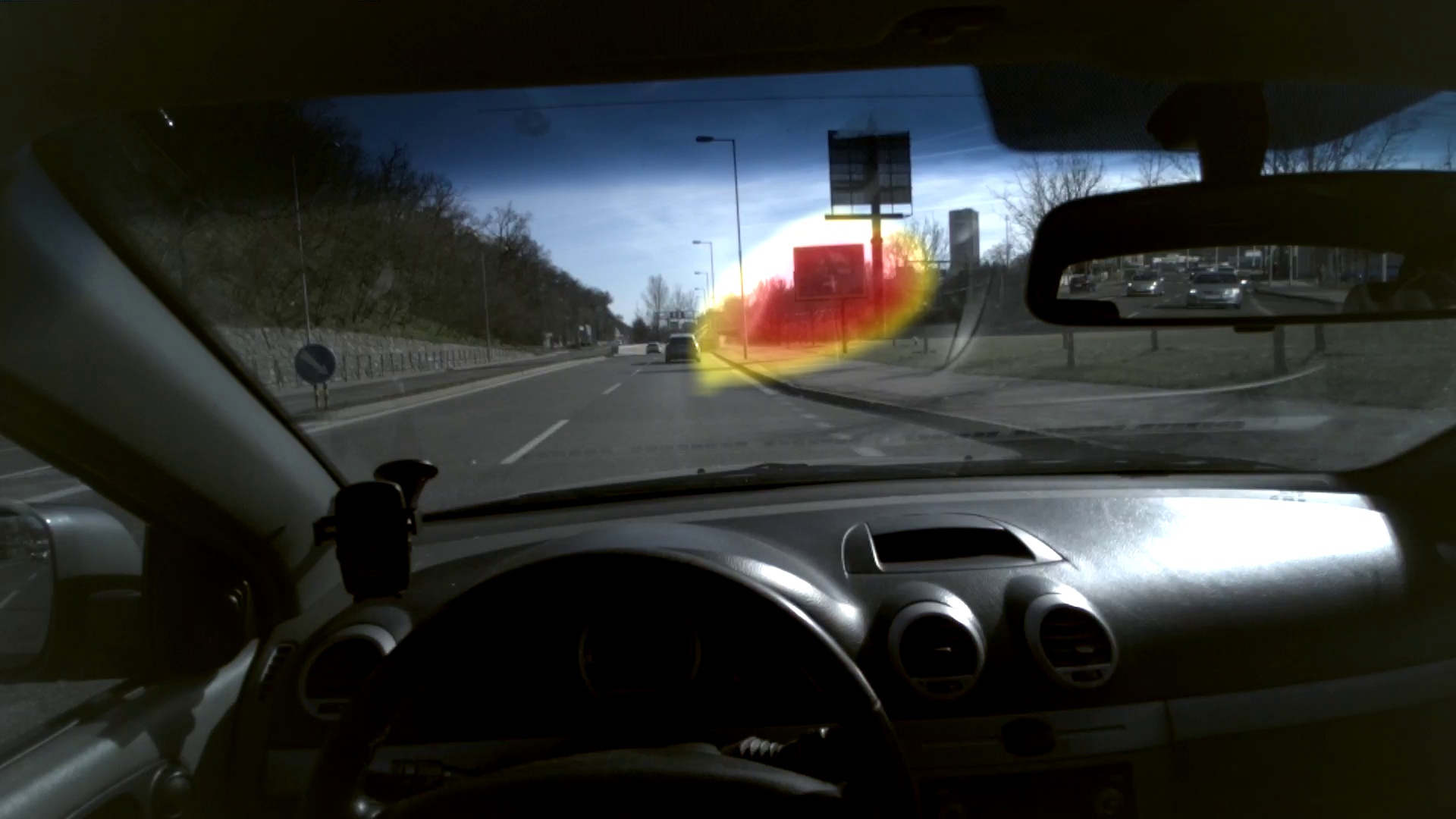}
    \caption{Fixations captured by Tobii Pro Glasses displayed on a frame from a video captured in the user study.}
    \label{fig:fix}
\end{figure}

To capture salient areas in the images and videos, we utilized the Tobii Pro Glasses 3, a wearable eye-tracking device providing robust and precise gaze data collection, all while allowing users unrestricted mobility. These glasses feature 16 illuminators and 4 eye cameras integrated into scratch-resistant lenses, ensuring optimal eye-tracking performance and an unobstructed view for the wearer. Additionally, the scene camera has a wide field of view (106° H: 95°, V: 63°). An example of the fixations captured by Tobii Pro Glasses displayed on the frame from a video captured in the user study can be seen in Fig. \ref{fig:fix}

\section{Billboard tracking}
To achieve information on the saliency of different billboards in the driving scenario it is necessary to analyse them not only as individual frames but as tracked instances. We decided to test YOLOv8large \cite{RangeKing23} in combination with different trackers based on the SORT method \cite{Brostrom23}.
\subsection{Detector}
\begin{figure}
    \centering
    \includegraphics[width=0.48\textwidth]{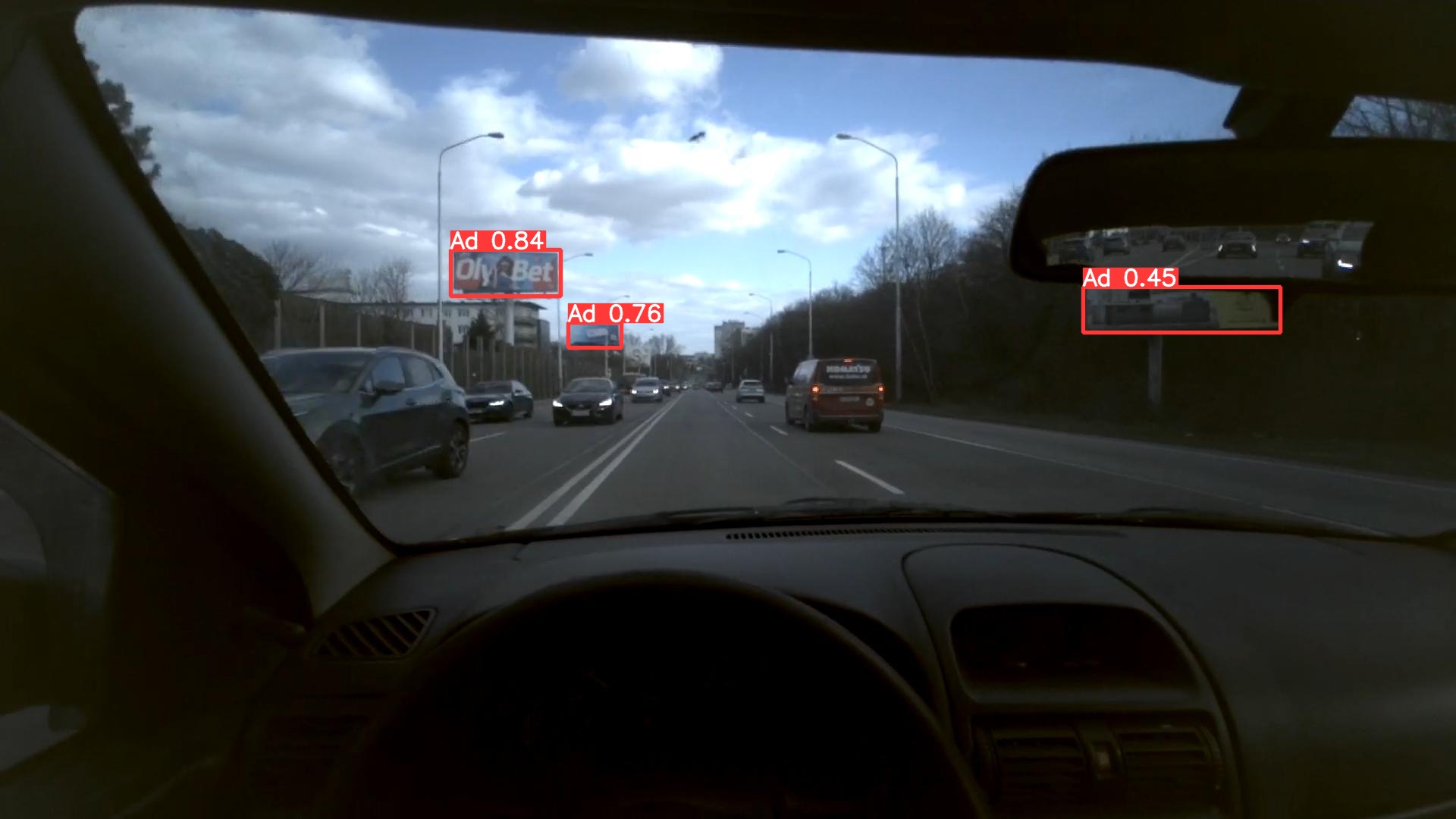}
    \caption{Frame from our BillboardLamac dataset with 3 billboards detected by the Yolov8large model \cite{RangeKing23} trained on a reduced version of the Mapillary Vistas 
    dataset \cite{Neuhold17}.}.
    \label{fig:det}
\end{figure}

The YOLOv8large \cite{RangeKing23} billboard detector was trained on Mappilary Vistas Dataset \cite{Neuhold17}.
Most of the images in Mapillary Vistas are captured from the interior or rooftop of vehicles, showcasing traffic within urban areas with numerous buildings, sidewalks, traffic signs, and various road objects, including advertising surfaces. We excluded images without advertisements, resulting in a reduced dataset of approximately 20,000 images. During the initial training, we noticed higher error rates specifically for smaller labeled advertisements. The smallest ads were labeled at such a distance that it would be uncertain whether the driver was looking at the advertisement or another object. Therefore, we created a second version of the dataset that only included images where the advertising surface accounted for at least 0.1\% of the image, resulting in almost half the number of images compared to the original dataset. To evaluate the detector we used the standard mAP50 metric. We have evaluated the trained YOLOv8large on both Mapilary Vistas and our dataset BillboardLamac and achieved \textbf{0.503 mAP50} and\textbf{ 0.656 mAP50} scores respectively. The higher score on BillboardLamac was gained due to less challenging billboards (in-occluded, bigger, etc.) in our dataset. Billboard's bounding boxes with corresponding confidences acquired by YOLOv8large on the BillboardLamac dataset can be seen in Fig. \ref{fig:det}. 

\subsection{Tracking}
\begin{figure}
    \centering
    \includegraphics[width=0.48\textwidth]{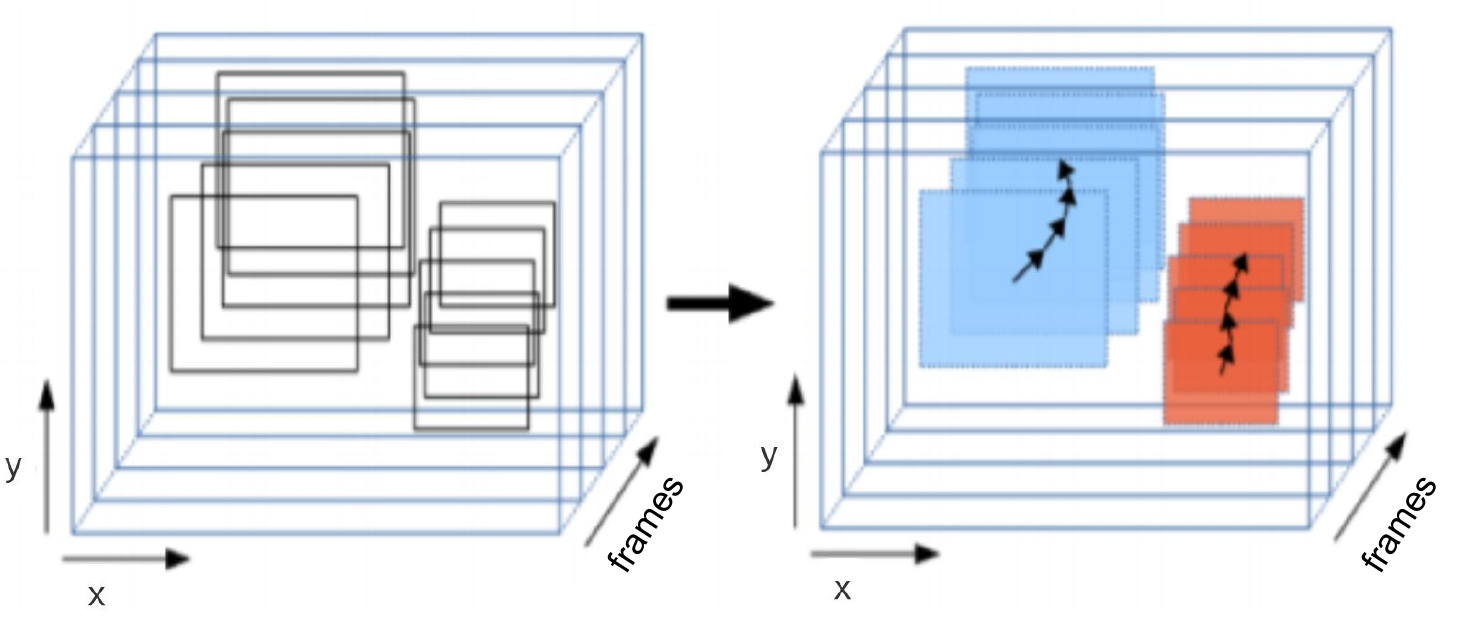}
    \caption{Tracking by detection. Objects are first detected via bounding boxes which are then associated into separate tracks.}
    \label{fig:track}
\end{figure}

To track billboards, we have tested the methods \mbox{OC-SORT}~\cite{Cao23}, Deep OC-SORT~\cite{Maggiolino23}, StrongSORT~\cite{Du23}, BoTSORT~\cite{Aharon22}, and ByteTrack~\cite{Zhang22}. All methods are based on the SORT (Simple Online and Realtime Tracking) method, which employs the Kalman filter and Hungarian algorithm for object tracking~\cite{Bewley16}. SORT is a tracking-by-detection framework for the problem of multiple object tracking (MOT) where objects are detected in each frame and represented as bounding boxes and the tracker associates them (see Fig. \ref{fig:track}). 
In our approach, the input video is processed frame by frame, and detections are obtained using the trained YOLOv8 model and associated with references. Multiple objects and classes can be tracked simultaneously. As shown in Fig. \ref{fig:track}, the output of the tracking method consists of detections with assigned references, describing the trajectory of the tracked object.

To evaluate object tracking we decided to use the HOTA (Higher Order Tracking Accuracy) metric~\cite{Luiten20}, which consists of smaller components that assess three aspects of tracking: localization, detection, and association. This formulation ensures that both detection and association, unlike in many other tracking metrics, are evenly balanced, and the final score lies somewhere between them. The results for each method are recorded in Table \ref{table:hota}.

\begin{table*}[ht]
\centering
\caption{Comparison of the evaluated tracking methods. We report metrics proposed in \cite{Luiten20}.}
\label{table:hota}
\begin{tabular}{|l l l l l l l l l|}
 \hline
Method & HOTA & DetPr & DetRe & DetA & AssPr & AssRe & AssA & Loc \\ [0.5ex]
 \hline
OC-SORT \cite{Cao23} & 36.8  &  39.5  &  34.3  &  53.2  &   58.2  &  36.3   &  \textbf{85.1}  &  89.1 \\ [0.1ex]
DeepOC-SORT \cite{Maggiolino23} & 35.8 &   38.9  &  33.0  &  54.4  &  55.6  &  34.8  &  84.7  &  89.1 \\ [0.1ex]
StrongSORT \cite{Du23} & \textbf{38.5}  &  \textbf{41.4}  &  \textbf{35.8}  &  \textbf{54.6} &   60.6   &   \textbf{42.7}   & 72.9   & 88.7 \\ [0.1ex]
BotSORT \cite{Aharon22} & 37.0  &  39.5  &  34.7  &  53.0  &  58.4  &  37.5 &   82.6  &  \textbf{89.2} \\ [0.1ex]
ByteTrack \cite{Zhang22} & 32.6  &  36.4  &  29.3  &  44.4  &  \textbf{62.8}  &  30.5  &  84.7  &  86.6 \\ [0.1ex]
 \hline
\end{tabular}
\end{table*}

The StrongSORT method performed the best in the overall HOTA metric and in most of the other metrics. However, in terms of association precision, the ByteTrack method achieved the highest scores, although it obtained lower values in almost all other metrics. For calculating the significance of billboards, the StrongSORT method proved to be the most effective, achieving the highest number of detections with relatively good reference assignments.

\subsection{Billboard Saliency}
If a billboard was detected in the frame, we analyzed the driver's gaze. The eyetracker captures up to two eye gaze fixation points per one video frame. In cases where no fixation point was recorded for a frame, we estimated it by extrapolating the positions of the two previous fixation points. If the information from the previous frame was also not available we did not consider the frame for further analysis. For frames where both a billboard was detected and a fixation point was recorded, we measured their overlap. This provided information on the number of frames where the driver looked at the billboard. By multiplying the number of frames by the duration of one frame, we obtained the gaze duration. Based on the viewing duration, we assessed the salience of the billboard for each driver individually and then calculated the final salience as median across the drivers.

A total of 145 billboards were identified along the chosen route. Table \ref{table:cat} presents the time range (represented by the number of frames with the billboard), the count of billboards in each category, and the average number of drivers who viewed the billboard for each category. The categorization was determined based on the outcomes of the StrongSORT tracking method.

On average, each billboard was viewed by three drivers, while 45 advertisements were not viewed by any driver. There were only two billboards viewed by all drivers.

\begin{table}[ht]
\centering
\caption{We split the detected billboards into 3 saliency categories based on the median gaze duration across the drivers. We also denote the average number of drivers who have looked at the billboards of the given cateogory.}
\label{table:cat}
\begin{tabular}{|l l c c|}
 \hline
 Category &	Gaze duration &	\# of billboards & Avg. \# of drivers \\ [0.5ex]
 \hline
None &	0 ms &	45 &	- \\ [0.1ex]
Short &	1-249 ms &	79 &	3 \\ [0.1ex]
Long &	250+ ms &	21 &	3 \\ [0.1ex]
 \hline
\end{tabular}
\end{table}

\section{Classification of Billboard significance}
We trained a Random Forest Classifier (RFC) to predict the significance of billboards. Data were divided into a training and a testing set. The training set contained 115 billboards and the test set had 30 billboards.

\subsection{Feature extraction}
We extracted a total of 7 features for each billboard to be used as input into the classifier. Each billboard was at most detected in 8 sequences. Not all of the billboards were detected in all videos, since billboards may have been hidden by cars or be present at such angles that they could not be seen when the driver was in the most distant lane of traffic. We calculated several features listed below for each billboard. To combine the data for each billboard from different driving sessions we have calculated the average over the sessions where the billboard was visible.

\subsubsection{Duration of visibility}
We included a feature indicating how many frames the ad has been visible based on the output of the tracker. We expect that the longer an ad has been visible, the more likely it is to be noticed by the driver.

\subsubsection{Advertisement localization}
Two features were computed from the localization of the advertisement. The first one determined in which part of the image the ad was located the longest. The image was divided into three sections: left, right, and center with ratios of 40:20:40. As a second feature, we also calculated the Euclidean distance of the center point of the ad from the center point of the image. We expect the position of the billboard to have a significant effect on the driver's attention.

\subsubsection{Billboard size}
Since the size of the advertisement increased as the vehicle approached the advertisement, we calculated the size of the advertisement as the average size over the entire sequence where the advertisement was visible. We also included a separate feature with the size calculated as the average for the 10 frames where the detected billboard bounding boxes were the largest.

\subsubsection{Saliency map}

\begin{figure}[t]
    \centering
    \includegraphics[width=0.48\textwidth]{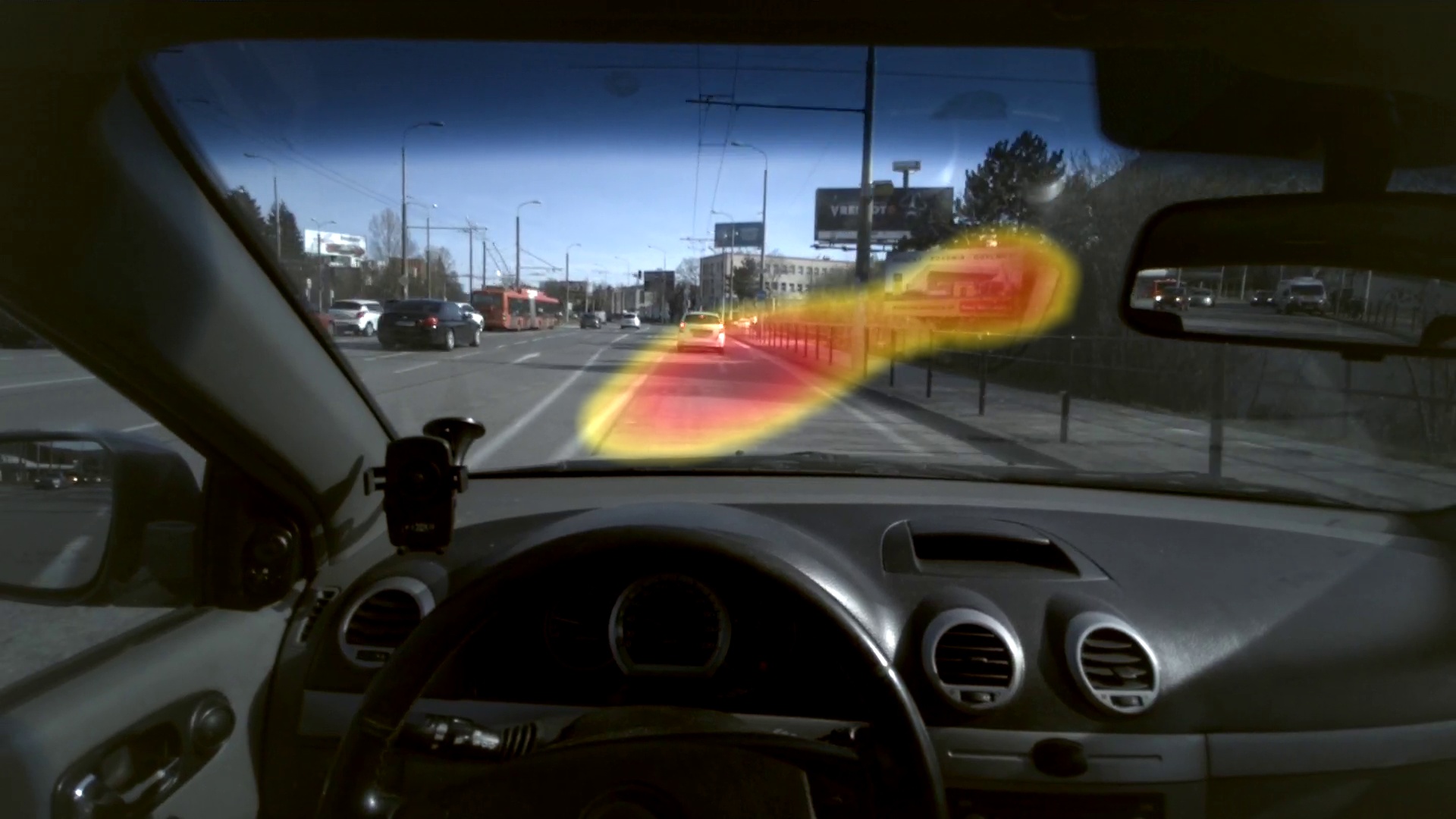}
    \caption{Example of overlaying the saliency map estimated by H2DS~\cite{hd2s} with a frame from the experiment.}
    \label{img:salmap}
\end{figure}

An existing neural network HD2S~\cite{hd2s} was used to generate saliency maps for each frame in the videos. The network outputs a map indicating the saliency of each frame. As depicted in Fig. \ref{img:salmap}, the saliency map effectively highlights the region containing the billboard. We evaluated the generated saliency map against the ground truth saliency obtained using the eyetracker with common metrics~\cite{Bylinskii18} resulting in AUC-Judd accuracy of $0.8964$, NSS of $1.7823$, SIM score of $0.3332$, and CC score of $0.4367$.

We included two features extracted from the saliency maps. One feature was calculated as the average of the saliency values within the detected area of the billboard and the second feature was calculated as the ratio of average saliency values in the billboard and the whole frame. Because of the many low or zero values in a frame that significantly reduce the average, all of the mentioned averages were calculated only for points with values greater than 50 (from range 0-255). To obtain these features for a given billboard we averaged them over the whole sequence of frames where the billboards were visible.

\subsection{Classification results}

We used a random forest classifier for classification. To optimize the training hyperparameters, we used 5-fold cross-validation. The trained classifier used 100 trees with a maximum depth of 2.

\begin{table}[]
\center
\renewcommand{\arraystretch}{1.7}
\begin{tabular}{cl|ccc|}
\cline{3-5}
\multicolumn{1}{l}{}                              &            & \multicolumn{3}{c|}{Predicted Class}                                \\ \cline{3-5} 
\multicolumn{1}{l}{}                              &            & \multicolumn{1}{c|}{0 ms} & \multicolumn{1}{c|}{0-250 ms} & 250+ ms \\ \hline
\multicolumn{1}{|c|}{\multirow{3}{*}{\rotatebox[origin=c]{90}{True Class}}} & 0 ms       & \multicolumn{1}{c|}{6}    & \multicolumn{1}{c|}{3}        & 0       \\ \cline{2-5} 
\multicolumn{1}{|c|}{}                            & 0 - 250 ms & \multicolumn{1}{c|}{1}    & \multicolumn{1}{c|}{15}       & 0       \\ \cline{2-5} 
\multicolumn{1}{|c|}{}                            & 250+ ms    & \multicolumn{1}{c|}{0}    & \multicolumn{1}{c|}{3}        & 1       \\ \hline
\end{tabular}
\vspace{0.5em}

\caption{The confusion matrix of the trained random forest classifier on the testing data.}
\label{tab:cm}
\end{table}

The resulting classifier accurately predicted the class for billboard saliency, and if incorrect, it was typically misclassified as a neighboring class. The overall classification performance achieved an accuracy of 75.8\% on the test set. The confusion matrix is shown in Table \ref{tab:cm}.

\begin{figure}[t]
    \centering

    \includegraphics[width=\linewidth]{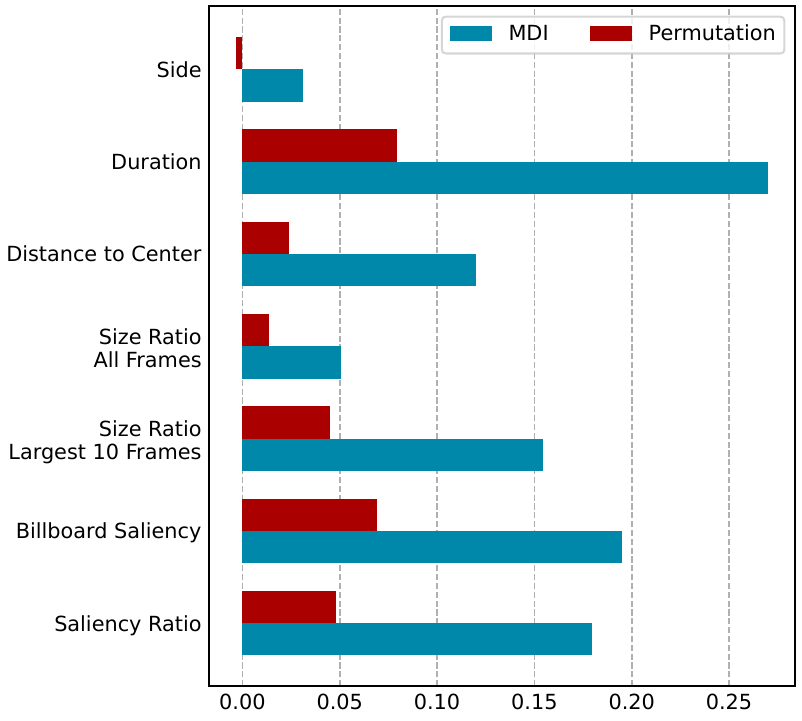}
    \caption{Feature importance for the trained random forest classifier calculated based on feature permutation (Permutation) and mean decrease in impurity (MDI).}
    \label{fig:importances}
\end{figure}

To analyse the impact of the features used for classification we performed feature importance analysis based on feature permutation and mean decrease in impurity. The resulting importance values can be seen in Fig. \ref{fig:importances}.

Among the various features considered, the duration of billboard visibility emerged as the most influential factor. This finding indicates that the length of time a billboard remains visible to drivers greatly impacts its perceived significance. Additionally, features derived from the saliency map, obtained using the HD2S method \cite{hd2s}, demonstrated a notable impact on billboard classification. This suggests that the visual prominence of billboards, as captured by saliency analysis, plays a crucial role in determining their effectiveness. Furthermore, we observed that the relative size of the billboard in the ten frames where it occupied the largest area also contributed significantly to the classification outcome.

By identifying these key features, our analysis provides valuable insights into the factors that contribute to the perceived significance of billboards. This information can guide future billboard design and placement strategies, enabling advertisers to optimize their campaigns for maximum impact and driver engagement.

\section{Conclusion}

We have presented a pipeline for evaluating the significance of roadside billboards in videos captured from a driver's perspective. Our approach does not require the use of eye-tracking devices during the actual pipeline execution.

Through the development and annotation of the BillboardLamac dataset, consisting of eight videos captured by drivers along a predefined path, we were able to train and evaluate our pipeline effectively. By evaluating various object tracking methods with the YOLOv8 detector, we achieved good performance in identifying billboard advertisements, with our best approach achieving a 38.5 HOTA score on BillboardLamac.

Moreover, we employed a random forest classifier to classify billboards into three distinct classes based on the length of driver fixations. This classifier demonstrated strong performance, achieving a test accuracy of 75.8\%. Notably, our analysis of the trained classifier revealed that the duration of billboard visibility, saliency features derived from a neural network, and the relative size of the billboard were the most influential factors when assessing billboard significance.

The findings from our feature importance analysis provide valuable insights for advertisers and marketers. Understanding the impact of factors such as visibility duration, visual saliency, and size allows for informed decisions regarding billboard design and placement strategies. By optimizing these factors, advertisers can enhance the effectiveness of their campaigns in capturing driver attention and engagement.

Overall, our pipeline offers a practical solution for evaluating the significance of roadside billboards, contributing to safer driving experiences and more effective marketing campaigns. Future work may involve expanding the dataset, incorporating additional features, and exploring other machine-learning techniques to further enhance the accuracy and applicability of the pipeline.

\section*{Acknowledgment}

This publication is the result of support under the Operational Program Integrated Infrastructure for the project: Advancing University Capacity and Competence in Research, Development and Innovation (ACCORD, ITMS2014+:313021X329), co-financed by the European Regional Development Fund. The work presented in this paper was carried out in the framework of the TERAIS project, a Horizon-Widera-2021 program of the European Union under the Grant agreement number 101079338.



\bibliographystyle{IEEEtran} 
\bibliography{references} 

\begin{thebibliography}{10}
\providecommand{\url}[1]{#1}
\csname url@samestyle\endcsname
\providecommand{\newblock}{\relax}
\providecommand{\bibinfo}[2]{#2}
\providecommand{\BIBentrySTDinterwordspacing}{\spaceskip=0pt\relax}
\providecommand{\BIBentryALTinterwordstretchfactor}{4}
\providecommand{\BIBentryALTinterwordspacing}{\spaceskip=\fontdimen2\font plus
\BIBentryALTinterwordstretchfactor\fontdimen3\font minus
  \fontdimen4\font\relax}
\providecommand{\BIBforeignlanguage}[2]{{%
\expandafter\ifx\csname l@#1\endcsname\relax
\typeout{** WARNING: IEEEtran.bst: No hyphenation pattern has been}%
\typeout{** loaded for the language `#1'. Using the pattern for}%
\typeout{** the default language instead.}%
\else
\language=\csname l@#1\endcsname
\fi
#2}}
\providecommand{\BIBdecl}{\relax}
\BIBdecl

\bibitem{Oviedo19}
\BIBentryALTinterwordspacing
O.~Oviedo-Trespalacios, V.~Truelove, B.~Watson, and J.~A. Hinton, ``The impact
  of road advertising signs on driver behaviour and implications for road
  safety: A critical systematic review,'' \emph{Transportation Research Part A:
  Policy and Practice}, vol. 122, pp. 85--98, 2019. [Online]. Available:
  \url{https://www.sciencedirect.com/science/article/pii/S0965856418310632}
\BIBentrySTDinterwordspacing

\bibitem{Beijer04}
D.~Beijer, A.~Smiley, and M.~Eizenman, ``Observed driver glance behavior at
  roadside advertising signs,'' \emph{Transportation Research Record}, vol.
  1899, no.~1, pp. 96--103, 2004.

\bibitem{Brome21}
R.~Brome, M.~Awad, and N.~M. Moacdieh, ``Roadside digital billboard
  advertisements: Effects of static, transitioning, and animated designs on
  drivers’ performance and attention,'' \emph{Transportation research part F:
  traffic psychology and behaviour}, vol.~83, pp. 226--237, 2021.

\bibitem{Yellappan16}
K.~Yellappan, Y.~Ghani, M.~Musa, M.~Siam, and C.~Tan, ``Exposure and perception
  on distraction towards roadside digital advertisements,'' 2016.

\bibitem{Smiley05}
A.~Smiley, B.~Persaud, G.~Bahar, C.~Mollett, C.~Lyon, T.~Smahel, and W.~L.
  Kelman, ``Traffic safety evaluation of video advertising signs,''
  \emph{Transportation research record}, vol. 1937, no.~1, pp. 105--112, 2005.

\bibitem{Costa19}
M.~Costa, L.~Bonetti, V.~Vignali, A.~Bichicchi, C.~Lantieri, and A.~Simone,
  ``Driver's visual attention to different categories of roadside advertising
  signs,'' \emph{Applied ergonomics}, vol.~78, pp. 127--136, 2019.

\bibitem{Crundall06}
D.~Crundall, E.~Van~Loon, and G.~Underwood, ``Attraction and distraction of
  attention with roadside advertisements,'' \emph{Accident Analysis {\&}
  Prevention}, vol.~38, no.~4, pp. 671--677, 2006.

\bibitem{Zalesinska18}
M.~Zalesinska, ``The impact of the luminance, size and location of led
  billboards on drivers’ visual performance—laboratory tests,''
  \emph{Accident Analysis {\&} Prevention}, vol. 117, pp. 439--448, 2018.

\bibitem{Mollu18}
K.~Mollu, J.~Cornu, K.~Brijs, A.~Pirdavani, and T.~Brijs, ``Driving simulator
  study on the influence of digital illuminated billboards near pedestrian
  crossings,'' \emph{Transportation research part F: traffic psychology and
  behaviour}, vol.~59, pp. 45--56, 2018.

\bibitem{Harasimczuk21}
J.~Harasimczuk, N.~E. Maliszewski, A.~Olejniczak-Serowiec, and A.~Tarnowski,
  ``Are longer advertising slogans more dangerous? the influence of the length
  of ad slogans on drivers’ attention and motor behavior,'' \emph{Current
  Psychology}, vol.~40, pp. 429--441, 2021.

\bibitem{Meuleners20}
L.~Meuleners, P.~Roberts, and M.~Fraser, ``Identifying the distracting aspects
  of electronic advertising billboards: A driving simulation study,''
  \emph{Accident Analysis {\&} Prevention}, vol. 145, p. 105710, 2020.

\bibitem{Maliszewski19}
N.~Maliszewski, A.~Olejniczak-Serowiec, and J.~Harasimczuk,
  ``\BIBforeignlanguage{eng}{Influence of sexual appeal in roadside advertising
  on drivers' attention and driving behavior},''
  \emph{\BIBforeignlanguage{eng}{PloS one}}, vol.~14, no.~5, p. e0216919, 2019.

\bibitem{Tarnowski17}
A.~Tarnowski, A.~Olejniczak-Serowiec, and A.~Marszalec, ``Roadside advertising
  and the distraction of driver’s attention,'' in \emph{MATEC Web of
  Conferences}, vol. 122.\hskip 1em plus 0.5em minus 0.4em\relax EDP Sciences,
  2017, p. 03010.

\bibitem{Marciano17}
H.~Marciano \emph{et~al.}, ``The effect of billboard design specifications on
  driving: a pilot study,'' \emph{Accident Analysis {\&} Prevention}, vol. 104,
  pp. 174--184, 2017.

\bibitem{Chan13}
M.~Chan and A.~Singhal, ``The emotional side of cognitive distraction:
  Implications for road safety,'' \emph{Accident Analysis {\&} Prevention},
  vol.~50, pp. 147--154, 2013.

\bibitem{RangeKing23}
\BIBentryALTinterwordspacing
G.~Jocher, A.~Chaurasia, and J.~Qiu, ``{YOLO by Ultralytics},'' Jan. 2023.
  [Online]. Available: \url{https://github.com/ultralytics/ultralytics}
\BIBentrySTDinterwordspacing

\bibitem{Brostrom23}
\BIBentryALTinterwordspacing
M.~Broström, ``{Real-time multi-object, segmentation and pose tracking using
  Yolov8 with DeepOCSORT and LightMBN}.'' [Online]. Available:
  \url{https://github.com/mikel-brostrom/yolov8\_tracking}
\BIBentrySTDinterwordspacing

\bibitem{Neuhold17}
G.~Neuhold, T.~Ollmann, S.~R. Bul{\`o}, and P.~Kontschieder, ``The mapillary
  vistas dataset for semantic understanding of street scenes,'' \emph{2017 IEEE
  International Conference on Computer Vision (ICCV)}, pp. 5000--5009, 2017.

\bibitem{Cao23}
J.~Cao, J.~Pang, X.~Weng, R.~Khirodkar, and K.~Kitani, ``Observation-centric
  sort: Rethinking sort for robust multi-object tracking,'' in
  \emph{Proceedings of the IEEE/CVF Conference on Computer Vision and Pattern
  Recognition}, 2023, pp. 9686--9696.

\bibitem{Maggiolino23}
G.~Maggiolino, A.~Ahmad, J.~Cao, and K.~Kitani, ``Deep oc-sort:
  Multi-pedestrian tracking by adaptive re-identification,'' \emph{arXiv
  preprint arXiv:2302.11813}, 2023.

\bibitem{Du23}
Y.~Du, Z.~Zhao, Y.~Song, Y.~Zhao, F.~Su, T.~Gong, and H.~Meng, ``Strongsort:
  Make deepsort great again,'' \emph{IEEE Transactions on Multimedia}, pp.
  1--14, 2023.

\bibitem{Aharon22}
N.~Aharon, R.~Orfaig, and B.-Z. Bobrovsky, ``Bot-sort: Robust associations
  multi-pedestrian tracking,'' \emph{arXiv preprint arXiv:2206.14651}, 2022.

\bibitem{Zhang22}
Y.~Zhang, P.~Sun, Y.~Jiang, D.~Yu, F.~Weng, Z.~Yuan, P.~Luo, W.~Liu, and
  X.~Wang, ``Bytetrack: Multi-object tracking by associating every detection
  box,'' in \emph{Computer Vision--ECCV 2022: 17th European Conference, Tel
  Aviv, Israel, October 23--27, 2022, Proceedings, Part XXII}.\hskip 1em plus
  0.5em minus 0.4em\relax Springer, 2022, pp. 1--21.

\bibitem{Bewley16}
A.~Bewley, Z.~Ge, L.~Ott, F.~Ramos, and B.~Upcroft, ``Simple online and
  realtime tracking,'' in \emph{2016 IEEE International Conference on Image
  Processing (ICIP)}, 2016, pp. 3464--3468.

\bibitem{Luiten20}
J.~Luiten, A.~Osep, P.~Dendorfer, P.~Torr, A.~Geiger, L.~Leal-Taix{\'e}, and
  B.~Leibe, ``Hota: A higher order metric for evaluating multi-object
  tracking,'' \emph{International journal of computer vision}, vol. 129, pp.
  548--578, 2021.

\bibitem{hd2s}
G.~Bellitto, F.~Proietto~Salanitri, S.~Palazzo, F.~Rundo, D.~Giordano, and
  C.~Spampinato, ``Hierarchical domain-adapted feature learning for video
  saliency prediction,'' \emph{International Journal of Computer Vision}, vol.
  129, pp. 3216--3232, 2021.

\bibitem{Bylinskii18}
Z.~Bylinskii, T.~Judd, A.~Oliva, A.~Torralba, and F.~Durand, ``What do
  different evaluation metrics tell us about saliency models?'' \emph{IEEE
  transactions on pattern analysis and machine intelligence}, vol.~41, no.~3,
  pp. 740--757, 2018.

\end{thebibliography}
\end{document}